# Graph-augmented Convolutional Networks on Drug-Drug Interactions Prediction

Yi Zhong[1,2,#], Xueyu Chen[3,#], Yu Zhao[1,2], Xiaoming Chen[1,2], Tingfang Gao[1,2], Zuquan Weng[1,2,*]

[1] College of Biological Science and Engineering, Fuzhou University, Fujian province, China;

[2] The Centre for Big Data Research in Burns and Trauma, College of Mathematics and Computer Science, Fuzhou University, Fujian province, China;

[3] School of Software, Shandong University, Shandong Province, China.

\* Corresponding author

   E-mail address: wengzq@fzu.edu.cn

[#] These authors contributed equally to this work.

**Abstract**

We propose an end-to-end model to predict drug-drug interactions (DDIs) by employing graph-augmented convolutional networks. And this is implemented by combining graph convolutional neural network with an attentive pooling network to extract structural relations between drug pairs and make DDI predictions. The experiment results suggest a desirable performance achieving ROC at 0.988, F1-score at 0.956, and AUPR at 0.986. Besides, the model can tell how the two DDI drugs interact structurally by varying colored atoms. And this may be helpful for drug design during drug discovery.

**Introduction**

Drug-drug interactions (DDIs) account for over 30% of all adverse drug reactions (ADRs) cases and often occur when co-medicate more than two drugs. More alarmingly, it stays a significant ADR-mediated morbidity every year[1], and this ramps up withdrawn-risks of a drug from the market and thus pulls a strong disincentive to drug development [2]. Though it is ideal for detecting all negative DDIs during clinical trials, DDIs-induced-ADRs cases are often reported at clinical uses and post-marketing surveillance, which pose a severe threat to public health. A study concerning the relationship between DDIs and the mortality rate of elderly hospitalized patients concludes that over 62.77% of patients present at least one DDI, and this may amount strictly to the death of these patients[3]. Besides, DDIs also expand the length of stay and cost of hospitalization[4]. Thus, predicting potential DDIs at an early stage, to the most possibility, makes DDIs-induced-ADRs evitable concerning the public expenses and health security. To date, two primary strategies, including medical trials and computational approaches, have been introduced to predict DDIs. The former focuses on biological and clinical trials to identify whether drug pairs share the same biochemical reactions or pathways [5] and induce ADRs.

However, it is highly constrained to identify all possible DDIs by conducting studies on patients who are on co-mediations during the clinical trials, which are also time and money consuming. Accordingly, to reduce costs and make broad possible predictions of DDIs, the latter has gained a growing attention in recent years.

By integrating biological and clinical trials and post-marketing surveillance (such as FDA Adverse Event Reporting System, FAERS), we can manually extract the feature vectors that represent DDIs drug pairs such as targets, phenotypes, side-effects, fingerprints, and so forth. Then using conventional supervised machine learning(ML) methods(like logistic regression(LR), random forest(RF), decision tree(DT), Naïve Bayes(NB),support vector machine(SVM)) to make predictions[6, 7].

Recently, Deep learning (DL) methods have witnessed rapid development in predicting DDIs. One reason is that DL can automatically extract features of drug pairs efficiently and effectively than manual works under different DL structures. For example, graph convolutional networks(GCNs) have been successfully implemented in graph learning, and thereupon it is potent to learn drug structures that are analogous to graphs[8]. Another is a great amount of DDIs data we can collect from knowledge-based databases such as DrugBank[9], the Kyoto Encyclopedia of Genes and Genomes (KEGG)[10]. With a large amount of data, DL has successfully excessed conventional ML in DDI predictions[11, 12].

In this paper, we propose a graph-augmented end-to-end neural network model that can predict DDIs (Figure 1). Specifically, we employ simplified molecular input line entry specification (SMILES) to extract graph topology (atoms stand for nodes, bonds for edges) for each drug separately that in each DDI pairs as inputs. Then a weight-shared graph convolutional neural network (GCNN) is applied to generate graph representations. After such GCNN layers, we embedded an attentive pooling network to calculate how two drugs interact. After a graph gather function, drug molecule representations in each pair are concatenated for DDIs representations, which are accepted by a deep neural network. Finally, a simple sigmoid function determines the final classification.

**Related works**

Various works related to predicting DDIs have been proposed, and they can be concluded as two types-knowledge embedding-based and similarity-based methods. The knowledge embedding-based model creates a DDI knowledge graph from knowledge databases, and implement embedding algorithms (e.g.RDF2Vec[13], PBG[14])[12] for fixed-vectors as network inputs, then applying deep learning network to make predictions. While drug similarity-based methods employed to integrate similarities such as drug structural, target, side effects similarity, etc. by similarity algorithms. Besides, to extract the structural relations between drugs, recent works have applied molecular fingerprints as inputs [15] to feed Siamese networks, which gained an excellent prediction performance. Besides, the latter cares about the structural similarity between DDI pairs, and it is beneficial for the beginning of drug discovery.

Learning a drug molecule also have two types. The one is to compute molecular SMILES to fixed-length fingerprint(such as PubChem fingerprint 881bits[16], Extended-Connectivity Fingerprints (ECFPs)[17]) by algorithms, and the other is to denote SMILES to a graph, then apply GCNN to learn features and these trained features are suitable for different tasks.

Attention mechanism has been widely used in vision, and natural language processing and several works have successfully brought this mechanism for biological mechanism explanation, especially to explain the relation between drug substructure and adverse reactions[18] and drug-target interactions[19]. And this currently has not been applied to explain the mechanism of DDIs.

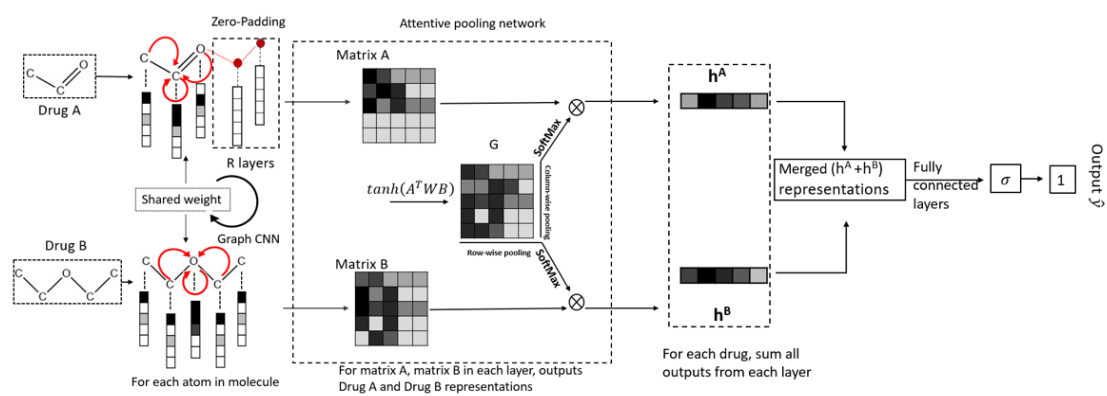

Figure 1 Overview of network architecture

**Figure 1** The inputs for the model are graph topology of drug pairs, for each drug in the dataset, the number of atoms is expanded to a max number 65. For each layer of GCNN, we adopt an attentive pooling network to make the alignment of two drugs and obtain attention vectors for each drug. After such layers, we respectively sum all attention vectors for each drug as one vector ($h^A$,$h^B$ in the figure). Then concatenated $h^A$,$h^B$ to feed fully-connected layers, and employ a sigmoid function to make final predictions.

## Material and methods

**Problem statement**

The inputs of the network are graph topology $G = (V, E)$[8] in which $V$ is a set of atoms, and $E$ is a set of bonds that connects two atoms. Given drug pair inputs, the model predicts $y \in \{0,1\}$, where $y = 1$ suggests DDI existence of drug pairs.

**DDI extraction and graph construction**

Our DDIs dataset is collected from the DrugBank database [20] Version 5.1.1, which is a golden standard DDI dataset covering 294,980 DDIs(positive) cross 2,286 approved drugs. Besides, we randomly pair an equally-sized negative dataset from the approved drugs group and exclude the pairs in the positive. Note that randomly paired drugs do not mean there are no DDIs. Though this may influence the prediction performance, the results are still acceptable[21].

We use RDKit to transform drug SMILES string to graph topology, and the transformation procedure partially refers to the code https://github.com/HIPS/neural-fingerprint. According to Duvenaud et.al. [22], each drug SMILES is converted to a graph where the nodes representing atoms and edges representing bonds respectively, and the features for nodes and edges are one-hot vectors which denote the types ( 62 atom types for nodes,6 bond types for edges) where nodes and edges belong to. The input size for the network is the number of nodes in a minibatch and the length of nodes features (nodes, node features), which means the input tensor difference for each minibatch. Considering the shape inconformity issue of drug pairs for attentive pooling network, we zero-padded nodes numbers (65 nodes for each drug in our model determined by the max atom number of a drug in our dataset) to where the drug molecule with the largest number of atoms in our dataset. Accordingly, the input tensor for our model is (batch size, nodes, node features). To avoid the influences of these zero-padding nodes in an attentive pooling network, we also created masks to ensure zero probability after the SoftMax function in attentive pooling.

**Graph Convolutional Neural Network (GCNN)**

We apply GCNN to transform the graph topology of a drug structure to the continuous hidden drug representations. The intuition is to view a drug molecule as an undirect graph G, with atoms as nodes and bonds as edges. Then the algorithm computes a molecule graph to a fixed size vector $r_g$ by aggregating the hidden representation $r_v^R$ of each atom in a molecule graph.

The $r_v^R$ for each atom in a molecule graph is calculated as follows:

$$f_v^{L+1} = r_v^L + \Sigma_{u \epsilon N(v)} r_u^L \quad (1)$$

$$r_v^{L+1} = \sigma(f_v^{L+1} H_L^{N(v)}) \quad (2)$$

Where L = 1 to R, R denotes numbers of layers(radii) for GCNN. $r_v^L$ is the hidden representation in the radius L. $N(v)$ is the neighbors of $v$. And $H_L^{N(v)}$ is a weight matrix. $\sigma$ in our model equals a relu function. Before propagation, $r_v^R$ is initialized by the *atom features* of $v$.

Then GCNN aggregates the hidden representation $r_v^R$ of each atom to final graph representations.

$$r_g = \sum_{v,L} softmax(r_v^L W^L) \quad (3)$$

Note that the equation (3) are omitted for attention implement and interpretability purpose. When initializing $r_v^R$, the atom features and the bond features are set as a 62 and 6-dimensional one-hot vector, respectively. Besides, in our network, we set radius R which determines the times the network will iterate as four.

**Siamese network**

Siamese network has been successfully adopted in image and text matching and visual tracking[23-25], Shortly speaking, the Siamese network is to evaluate the similarity of the two input samples by two weight shared sub-networks. Mapping DDI prediction, we take the view that drug pairs with DDIs may have a high-level structural relation. Accordingly, we add two weight-shared GCNN as sub-networks of Siamese, trying to learn molecule sub-structures that contribute to DDIs.

**Attentive pooling network**

The attention mechanism is one of the research hotspots in vision and natural language processing (NPL) field. Network with attention calculates the matching degree between input sequences and output vectors, and a high matching degree gains a high attention weight, which can selectively focus on the important spot of inputs, thus augment the performance of models. To figure out how drugs interact, we extend our model with an attentive pooling network[26], which is a two-way attention mechanism that builds a close connection between the two inputs pairs (question Q and answer A in Santos Cd et.al.). The algorithms apply over the outputs of CNN or LSTM to compute a soft alignment $G = tanh(Q^T W A)$, which conveys the interaction correlations scores between Q and A. Finally, applying a column-wise and row-wise max-pooling over *G* followed by a normalized operation SoftMax, the two-sides separately integrate mutual information.

For our model, given $P = tanh(A^T W B)$, where $A, B$ represent $A = M^{H_a \times L_a}$ and $B = M^{H_b \times L_b}$ Respectively. Where $H_a$ and $H_b$ are the hidden space dimensions of each drug in pairs, while $L_a$ and $L_b$ are the number of atoms of each drug in pairs. $W = M^{H_a \times H_b}$ is a trainable weight matrix.

**Model training**

We split our dataset into train and test dataset with radio 9:1. And perform the hyperparameter optimization through a 5-fold cross-validation test. We employed minibatch for training with Adam optimizer with learning rate 0.001; we hold the number of GCNN layers as four with the same units 50. As for the attentive pooling network, the Xavier initializer [27] is used to initialize weights. Besides, three fully connected layers are applied with the same units 100. The model are trained for 100 epochs with each epoch 200 steps. Except for the activation sigmoid for the final classification, all the activation functions are relu.

In training, the DDI classification model is conducted by minimizing cross-entropy loss function $L_i = -\sum_{i=1}^{n} y_i \log(\hat{y}_i)$ where $i = 1, \dots, n$ given a dataset $D = \{D1_i, D2_i, y_i\}$, $D1, D2$ are input drug pairs.

**Results**

**Analysis of DDI prediction**

| Model | ROC | F1-score | AUPR |
|---|---|---|---|
| IDNDDI[28] | 0.969 | — | — |
| Karim M.R et al.[12] | — | 0.92 | 0.940 |
| R. Celebi et al.[21] | 0.932 | 0.860 | — |
| Guy Shtar et al[29] | 0.991 | — | 0.960 |
| GCNN + Attentive pooling | 0.988 | 0.956 | 0.986 |

**Table 1** Comparison crosses the state-of-the-art DDI prediction models.

Table 1 shows the model performance of GCNN with Attention and the comparison with advanced DDI prediction models. From the table, we can notice that our model outperforms most of the models except for Guy Shtar et al. s' work when concerning ROC. Besides, the F1-score and AUPR score achieved 0.956 and 0.986, respectively, that overpass all the listed models.

**Case study for interpretability**

A key advantage of our model over baselines is the interpretability, where we can highlight the drug substructures that contribute to DDIs, and this, to some extent, helps explain the mechanism of DDIs incidences. During drug design, it is also helpful for experts to focus on the undesirable substructures of candidate chemicals. With the attention mechanism, we consider that the highest attention weights of nodes with their neighbors in drug pairs have the closest relation to DDI. Note that, the outputs of different layers for attentive pooling in GCNN contribute to different attention weights. To visualize, we compare attention weights cross each layer, and the attention weights with the highest one will be chosen for visualization.

Celecoxib (a kind of a nonsteroidal anti-inflammatory drug) and Mephenytoin (designed for treating refractory partial epilepsy) have been observed CYP2D6-associated drug-drug interactions, and this may lead to potential adverse effects. The hydroxylation of celecoxib inhibits the Cytochrome P450 2D6 (CYP2D6) (an enzyme that in humans is encoded by the CYP2D6 gene)[30], which result in the O-demethylation of dextromethorphan in human liver. To conduct this case study, we downloaded the two drug SMILES from DrugBank (Accession Number DB00482 and DB00532 respectively). Firstly, we predicted the interaction between the two drugs as positive with confidence 0.97, and by comparing the attention weights cross GCNN layers, attention weights from layer one is chosen for a highlight. We then used rdkit to visualize the highlighted atoms from the two drugs (Figure 2), and their weights color all atoms from high to low. We found that the highlighted atoms are correctly showed the sites where the two drugs interact according to the literature[30].

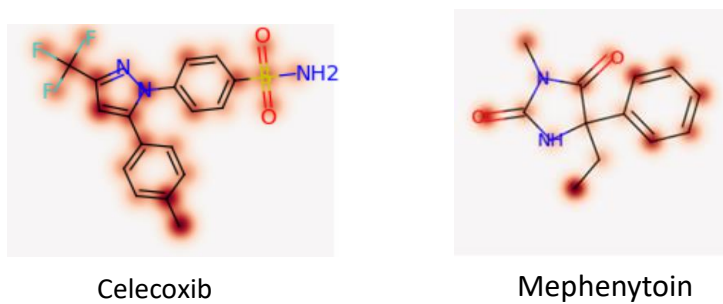

| Celecoxib | Mephenytoin |

**Figure 2** The highlighted atoms of the interaction drugs (Celecoxib and Mephenytoin).

## Conclusions

We have presented an end-to-end model to predict DDIs and the experimental results show that this model can achieve desirable prediction performance. And with the attention mechanism embedded in our model, the interaction mechanism of drugs can be interpreted by highlight atoms.

And this may help for drug discovery and predict the potential DDI pairs.

## Acknowledgment

This work was supported by National Natural Science Foundation of China (No. 81971837), Funds of Joint Plan for Health Education in Fujian (No. WKJ2016-2-25), and Start-up Funds of Fuzhou University, Fujian, China (No. XRC-19015).

## References

1.  Iyer SV, Rave H, Paea LP, Anna BM, Shah NH: **Mining clinical text for signals of adverse drug-drug interactions**. *Journal of the American Medical Informatics Association* 1900, **21**(2):353-362.

2.  Boissonnat P, Lorgeril MD, Perroux V, Salen P, Batt AM, Barthelemy JC, Brouard R, Serres E, Delaye J: **A drug interaction study between ticlopidine and cyclosporin in heart transplant recipients**. *European Journal of Clinical Pharmacology* 1997, **53**(1):39-45.

3.  Rosas-Carrasco Ó, García-Peña C, Sánchez-García S, Vargas-Alarcón G, Gutiérrez-Robledo LM, Juárez-Cedillo T: **The relationship between potential drug-drug interactions and mortality rate of elderly hospitalized patients**. *Revista de Investigación Clínica* 2011, **63**(6):564-573.

4.  Moura CS, Acurcio FA, Belo NO: **Drug-drug interactions associated with length of stay and cost of hospitalization**. *Journal of pharmacy & pharmaceutical sciences : a publication of the Canadian Society for Pharmaceutical Sciences, Societe canadienne des sciences pharmaceutiques* 2008, **12**(3):266.

5.  Balian JD, Rahman A: **Metabolic Drug-Drug Interactions: Perspective from FDA Medical and Clinical Pharmacology Reviewers**. *Advances in Pharmacology* 1997, **43**:231.

6.  Group BP: **Journal of the American Medical Informatics Association**. *Journal of the American Medical Informatics Association* 2009, **22**(2).

7.  Heng L, Ping Z, Hui H, Jialiang H, Emily K, Leming S, Lin H, Lun Y: **DDI-CPI, a server that predicts drug-drug interactions through implementing the chemical-protein interactome**. *Nucleic Acids Research* 2014, **42**(W1):W46-W52.

8.  Altae-Tran H, Ramsundar B, Pappu AS, Pande V: **Low Data Drug Discovery with One-Shot Learning**. *Acs Central Science* 2016, **3**(4):283.

9.  Wishart DS, Knox C, An CG, Shrivastava S, Hassanali M, Stothard P, Zhan C, Woolsey J: **DrugBank: a comprehensive resource for in silico drug discovery and exploration**. *Nucleic Acids Research* 2006, **34**(Database issue):668-672.

10. Kanehisa M, Goto S: **KEGG: kyoto encyclopedia of genes and genomes**. *Nucleic Acids Research* 2000, **27**(1):29-34.

11. Ryu JY, Kim HU, Lee SY: **Deep learning improves prediction of drug-drug and drug-food interactions**. *Proceedings of the National Academy of Sciences of the United States of America* 2018, **115**(18):201803294.

12. Karim MR, Cochez M, Jares JB, Uddin M, Beyan O, Decker S: **Drug-Drug Interaction Prediction Based on Knowledge Graph Embeddings and Convolutional-LSTM**

**Network**. In: *Proceedings of the 10th ACM International Conference on Bioinformatics, Computational Biology and Health Informatics: 2019*. ACM: 113-123.

13. Ristoski P, Rosati J, Di Noia T, De Leone R, Paulheim H: **RDF2Vec: RDF graph embeddings and their applications**. *Semantic Web* 2019, **10**(4):721-752.

14. Lerer A, Wu L, Shen J, Lacroix T, Wehrstedt L, Bose A, Peysakhovich A: **PyTorch-BigGraph: A Large-scale Graph Embedding System**. *arXiv preprint arXiv:190312287* 2019.

15. Ryu JY, Kim HU, Lee SY: **Deep learning improves prediction of drug–drug and drug–food interactions**. *Proceedings of the National Academy of Sciences* 2018, **115**(18):E4304-E4311.

16. Cereto-Massagué A, Ojeda MJ, Valls C, Mulero M, Garcia-Vallvé S, Pujadas G: **Molecular fingerprint similarity search in virtual screening**. *Methods* 2015, **71**:58-63.

17. Rogers D, Hahn M: **Extended-connectivity fingerprints**. *Journal of chemical information and modeling* 2010, **50**(5):742-754.

18. Dey S, Luo H, Fokoue A, Hu J, Zhang P: **Predicting adverse drug reactions through interpretable deep learning framework**. *BMC bioinformatics* 2018, **19**(21):476.

19. Gao KY, Fokoue A, Luo H, Iyengar A, Dey S, Zhang P: **Interpretable Drug Target Prediction Using Deep Neural Representation**. In: *IJCAI: 2018*. 3371-3377.

20. Wishart DS, Feunang YD, Guo AC, Lo EJ, Marcu A, Grant JR, Sajed T, Johnson D, Li C, Sayeeda Z: **DrugBank 5.0: a major update to the DrugBank database for 2018**. *Nucleic acids research* 2017, **46**(D1):D1074-D1082.

21. Celebi R, Yasar E, Uyar H, Gumus O, Dikenelli O, Dumontier M: **Evaluation of Knowledge Graph Embedding Approaches for Drug-Drug Interaction Prediction using Linked Open Data**. 2018.

22. Duvenaud DK, Maclaurin D, Iparraguirre J, Bombarell R, Hirzel T, Aspuru-Guzik A, Adams RP: **Convolutional networks on graphs for learning molecular fingerprints**. In: *Advances in neural information processing systems: 2015*. 2224-2232.

23. Jiang C, Xiao J, Xie Y, Tillo T, Huang K: **Siamese network ensemble for visual tracking**. *Neurocomputing* 2018, **99**(99):S0925231217316946.

24. Pei W, Tax DMJ, Laurens VDM: **Modeling Time Series Similarity with Siamese Recurrent Networks**. 2016.

25. Melekhov I, Kannala J, Rahtu E: **Siamese network features for image matching**. In: *International Conference on Pattern Recognition: 2017*.

26. Santos Cd, Tan M, Xiang B, Zhou B: **Attentive pooling networks**. *arXiv preprint arXiv:160203609* 2016.

27. Glorot X, Bengio Y: **Understanding the difficulty of training deep feedforward neural networks**. In: *Proceedings of the thirteenth international conference on artificial intelligence and statistics: 2010*. 249-256.

28. Yan C, Duan G, Zhang Y, Wu F-X, Pan Y, Wang J: **IDNDDI: An Integrated Drug Similarity Network Method for Predicting Drug-Drug Interactions**. In: *International Symposium on Bioinformatics Research and Applications: 2019*. Springer: 89-99.

29. Shtar G, Rokach L, Shapira B: **Detecting drug-drug interactions using artificial neural networks and classic graph similarity measures**. *arXiv preprint arXiv:190304571* 2019.

30. Siu YA, Hao M-H, Dixit V, Lai WG: **Celecoxib is a substrate of CYP2D6: impact on**

celecoxib metabolism in individuals with CYP2C9* 3 variants. *Drug metabolism and pharmacokinetics* 2018, **33**(5):219-227.